\documentclass[conference]{IEEEtran}
\IEEEoverridecommandlockouts
\usepackage{multirow}
\usepackage{cite}
\usepackage{amsmath,amssymb,amsfonts}
\usepackage{algorithmic}
\usepackage{graphicx}
\usepackage{textcomp}
\usepackage{xcolor}
\usepackage[a4paper, total={184mm,239mm}]{geometry}
\def\BibTeX{{\rm B\kern-.05em{\sc i\kern-.025em b}\kern-.08em
    T\kern-.1667em\lower.7ex\hbox{E}\kern-.125emX}}

\usepackage{subcaption}

\usepackage{mathrsfs}  
\usepackage{bm}

\newcommand{\equationskip}{%
\setlength{\abovedisplayskip}{3pt}
\setlength{\belowdisplayskip}{3pt}
\setlength{\abovedisplayshortskip}{2pt}
\setlength{\belowdisplayshortskip}{2pt}
}

\usepackage{enumitem}

\setenumerate[1]{itemsep=0.5pt,partopsep=0pt,parsep=\parskip, topsep=2pt, leftmargin=12pt,}

\newcommand*{\Toolname}{FitAct}

\begin{document}

\title{\Toolname: Error Resilient Deep Neural Networks via  \underline{Fi}ne-Grained Post-\underline{T}rainable \underline{Act}ivation Functions
% \thanks{}
}
\author{\IEEEauthorblockN{Behnam Ghavami, Mani Sadati, Zhenman Fang, Lesley Shannon}
\IEEEauthorblockA{Simon~Fraser~University, Burnaby, BC, Canada\\
Emails: \{behnam\_ghavami, zhenman, lesley\_shannon\}@sfu.ca}}
% \and
% \and
% \and
% \and
% \and

\maketitle
\begin{abstract}
Deep neural networks (DNNs) are increasingly being deployed in safety-critical systems such as personal healthcare devices and self-driving cars. In such DNN-based systems, error resilience is a top priority since faults in DNN inference could lead to mispredictions and safety hazards. For latency-critical DNN inference on resource-constrained edge devices, it is nontrivial to apply conventional redundancy-based fault tolerance techniques. 
%where high . On the other hand, implementation of highly error resilient DNN-based systems is difficult since .
%Network architecture modification is emerging as a promising solution for achieving efficient error resilience in DNNs, particularly in resource constraint devices.

In this paper, we propose \Toolname, a low-cost approach to enhance the error resilience of DNNs by deploying fine-grained post-trainable activation functions. 
The main idea is to precisely bound the activation value of each individual neuron via neuron-wise bounded activation functions, so that it could prevent the fault propagation in the network. 
%Different to prior studies that use a global bound value for all neurons in a layer, we propose to provide a fine-grained bound value for each neuron
To avoid complex DNN model re-training, we propose to decouple the accuracy training and resilience training, and develop a lightweight post-training phase to learn these activation functions with precise bound values.
Experimental results on widely used DNN models such as AlexNet, VGG16, and ResNet50 demonstrate that \Toolname~outperform state-of-the-art studies such as Clip-Act and Ranger in enhancing the DNN error resilience for a wide range of fault rates, while adding manageable runtime and memory space overheads.

% The way it works is to adjust some unique parameters via the post-training stage in order to precisely bound each individual neuron in a DNN model so that it prevents the neuron's values from increasing deviation because of the faults on memory where the network's parameters are stored. Detailed experiments on some popular DNNs, AlexNet, VGG16 and ResNet50, demonstrate that \Toolname~is highly effective in enhancing the error resilience with a negligible extra computation overhead.
\end{abstract}

% \keywords{DNN, Error Resilience, Hardware Faults, Activation Function}

% \begin{abstract}
% This document is a model and instructions for \LaTeX.
% This and the IEEEtran.cls file define the components of your paper [title, text, heads, etc.]. *CRITICAL: Do Not Use Symbols, Special Characters, Footnotes, 
% or Math in Paper Title or Abstract.
% \end{abstract}

% \begin{IEEEkeywords}
% DNN, Error Resilience, Hardware Faults, Activation Function, ReLU.
% \end{IEEEkeywords}

%%%%%%%%%%%%%%

% \begin{document}

% \title{\ToolName: Error Resilience Design of Deep Neural Networks via a  Fine-grained Trainable Activation Function}

% \input{abstract}
% \maketitle
%

\section{Introduction}\label{intro}
% Application of DNN in safety-critical domain
% Problems with traditional fault tolerance methods
%

Recently, Deep Neural Networks (DNNs) have brought some promising results into safety-critical applications such as cyber-physical systems, personal healthcare devices \cite{miotto2018deep} and self-driving vehicles \cite{grigorescu2020survey}.
However, to realize successful deployment of DNNs in real-world safety-critical systems, in addition to their inference accuracy and latency, one also needs to consider the error resilience of DNNs under environmental noises and hardware memory faults \cite{shafique2020robust}\cite{mittal2020survey}. In this paper, the error resilience of a DNN is defined as its ability to maintain the model accuracy in the presence of random faults; under random faults, the higher the model accuracy, the better the model resilience.

% the challenges on the design of error resilient DNN accelerators must be addressed. In other words, error resiliency is a major concern for DNN accelerators in safety-critical applications, in addition to efficiency.  Environmental noises and hardware memory faults can compromise the inherent resiliency of DNN models during inference \cite{shafique2020robust}. 

Traditionally, the error resilience of a system is usually enhanced by redundancy-based fault tolerance techniques, where one of more dimensions of hardware, software, data, and time are duplicated to tolerate the faults \cite{mahmoud2020hardnn,li2017understanding,libano2018selective,sabih2021fault,hari2021making,dos2018analyzing,kosaian2021arithmetic}. 
Such techniques usually entail significant cost, latency and/or resource overheads, which are often not suitable for latency-critical DNN inference on resource-constrained edge devices. More related work will be discussed in Section~\ref{sec:related}.

%DNN-based applications typically require a large amount of computational resources, and thus cannot be adequately supported by traditional redundant-based fault tolerance techniques, where protecting computing systems from errors using hardware/software/time/data redundancies frequently entail significant cost, latency and power consumption overheads. These overheads, in particular, contrast with the low-cost, low-power, and low-latency requirements of edge-based DNN inference. Hence, traditional fault tolerance techniques are not effective to be deployed in real-time DNN edge devices. 
%As a result, an efficient strategy to tolerate DNN accelerators against errors without incurring considerable overheads is urgently needed.

To overcome the above challenge, one appealing direction is to enhance the error resilience of DNNs via simple network architecture modifications \cite{dias2010ftset,schorn2019efficient,nwankpa2018activation,liew2016bounded,chen2020ranger,hong2019terminal,hoang2020ft,ali2020erdnn}. More specifically, one promising solution is to bound the activation values of neurons in the network to help prevent the fault propagation by making simple changes to the activation functions in a DNN model \cite{chen2020ranger,hong2019terminal,hoang2020ft}. However, as we will present in Section~\ref{subsec:BoundRELIssu}, we observe that prior studies  \cite{chen2020ranger,hong2019terminal,hoang2020ft} use a global bound value for all neurons in a network layer and become ineffective when the fault rate becomes $10^{-6}$ and higher. The reason is that the (normal) maximum activation values of neurons vary wildly. 

Inspired by the above observation, in this paper, we propose to employ a fine-grained neuron-wise activation function in the DNN model, which has an adjustable upper-bound for each neuron to better enhance the model error resilience. One challenge behind this simple idea is that it introduces a huge number of different bound values for users to figure out. To address this challenge, in Section~\ref{sec:proposedAF}, we propose a trainable fine-grained activation function, called FitReLU, based on the most widely used ReLU (Rectified Linear Unit) \cite{nwankpa2018activation} activation function. 
To avoid retraining a DNN model from scratch and complicating the training process, in Section~\ref{sec:proposedDNN}, we propose to decouple the accuracy training (conventional training) and resilience training (post-training), and develop a two-stage framework called \Toolname~to enhance the model resilience. In \Toolname, we develop a lightweight post-training phase to learn those fine-grained activation functions with precise bound values.

%In this paper, we propose an efficient technique, named as \Toolname, to improve the error resilience of DNNs based on a new architecture modification strategy. Our main idea is to employ a fine-grained trainable activation function that has an adjustable upper-bound for each neuron and replace it with the popular activation function (e.g ReLU \cite{nwankpa2018activation}) in the network. The advantage of locally introducing the parameters for bounding the activation over its global counterparts \cite{chen2020ranger,hong2019terminal,hoang2020ft} is that, it gives additional degrees of freedom to every neuron in all the hidden layers, which in turn can be used to get more resiliency. 
%In order to keep the DNN training process from becoming overly complicated, we also introduce a post-training stage to the network design flow to adjust the activation bounds of the network. %So the goal of the post train parameter adjustment algorithm is to find the best upper-bound values for each neuron so that the model still gives same baseline accuracy and gives a higher robustness.

In our experiments, we perform fault simulations on three widely used DNN models, AlexNet, VGG16, and ResNet50, on two datasets, CIFAR-10 and CIFAR-100, under various fault rates ranging from $10^{-7}$ to $3\times10^{-5}$. Compared to state-of-the-art studies Clip-Act \cite{hoang2020ft} and Ranger \cite{chen2020ranger} that use a global bound for all neurons in a network layer, \Toolname~achieves better error resilience, especially at higher fault rates that go beyond $10^{-6}$. For example, at the fault rate of $3\times10^{-6}$, \Toolname~can still achieve a high accuracy of 84.81\% and 90.28\% for ResNet50 and VGG16 models on CIFAR-10 dataset, while Clip-Act can only achieve an accuracy of 52.47\% and 61.63\%, and Ranger almost provides no protection. 
In general, compared to the original DNN model with regular ReLU activation function, \Toolname~adds less than 7\% extra runtime overhead for the post-training step, less than 12\% extra runtime overhead and less than 6\% memory space overhead for the inference step.

\section{Related Work}\label{sec:related}

%The assessment and enhancement of DNNs' resilience against faults has been an interesting subject over the past decade \cite{mittal2020survey} and has got increasing attention recently, due to the advance of DNN deployment in safety-critical applications. 
In general, the proposed techniques to enhance the error resilience of DNNs can be classified into the following categories. %\cite{reagen2018ares,li2017understanding,li2018tensorfi,chen2019binfi,he2020fidelity}.
%  \cite{reagen2018ares}\cite{li2017understanding} \cite{li2018tensorfi}\cite{chen2019binfi}\cite{he2020fidelity}An extensive literature review has been given by .

\subsection{Traditional Hardware Redundancy-based Fault Mitigation}
Redundancy-based fault detection and mitigation techniques are commonly used for mitigating faults in machine learning hardware. For example, Tesla’s self-driving cars used expensive Dual Modular Redundancy (DMR) to mitigate the impact of faults \cite{motors2019full}.
A more efficient technique is selective node hardening, which often suffers from limited fault coverage though.
For example, Mahmoud et al. \cite{mahmoud2020hardnn} used statistical fault study to identify vulnerable regions in DNNs, and selectively duplicate the vulnerable computations for fault detection.  
Li et al. \cite{li2017understanding} leveraged the value spikes in the neural responses as the symptoms for fault detection. However, program re-execution is required to restore the correct output.
A relaxed version of Triple Modular Redundancy (TMR) was deployed for FPGA-based DNN accelerators in \cite{libano2018selective}. Sabih et. al. \cite{sabih2021fault} utilized explainable deep learning techniques to make DNNs more robust by identifying vulnerable parts and selectively protecting them using Error Correction Codes (ECC) and TMR.

Although these approaches offer improved resilience against faults, they have high time or resource overheads and are not preferable for latency-critical DNN inference on edge devices.

\subsection{Traditional Algorithm Redundancy-based Fault Tolerance}
%\zf{Please elaborate more about ABFT, how does it detect and tolerate faults?}
Algorithm-Based Fault Tolerance (ABFT) has also been proposed to detect (and tolerate) faults in some particular layers of DNNs (e.g., convolutional layer \cite{hari2021making} or fully-connected layer \cite{dos2018analyzing}) by inserting checksums for layer operations %of those layers 
in the algorithm. 
To compensate the execution time overhead, Kosaian et. al. \cite{kosaian2021arithmetic} investigated thread-level ABFT schemes for GPUs that exploit an adaptive arithmetic intensity guided approach.

However, these techniques do not protect DNNs against faults occurring in other layers of a DNN or fault propagation into multiple neurons (in multiple layers). In addition, they also have remarkable time overhead.% and thus are not recommended for DNN deployment on edge devices.

\subsection{Fault-Aware Training}
Some studies account for the fault patterns of the inference accelerator during the training itself \cite{kim2018matic}. For example, a research in  \cite{jia2018calibrating} performed retraining to reduce the impact of errors in analog domains. 
Common regularizing techniques via training process, such as dropout, may improve the general error resilience of models  as well \cite{mhamdi2017neurons}. Zahid et al. \cite{zahid2020fat} introduced a fault-aware training that includes permanent fault modeling during training of FPGA-based DNN accelerators.

However, these approaches complicate the training process, since fault injections need to be performed by embedding inference hardware in the training procedure or through exhaustive fault simulations,  both of which are impractical.

\subsection{Modifications of the Network Architecture}

Another common approach is to modify the neural network architecture to increase its error resilience, which can be done either during training or outside training. 
For example, Dias et al. \cite{dias2010ftset} suggested a resilience optimization procedure by changing the architecture in order to diminish the impact of the faults. However, they used exhaustive simulations to determine critically values, which is impractical for large DNNs. 
Schorn et al. \cite{schorn2019efficient} introduced a resilience enhancement procedure by the replication of critical layers and features to achieve a more homogeneous resilience distribution within the network. Nevertheless, no automated design flow was introduced to jointly optimize the error resilience and accuracy of DNNs.
%\cite{panigrahi2019effect, nwankpa2018activation}

Recently, Hong et al. \cite{hong2019terminal} introduced an efficient approach to mitigate memory faults in DNNs by modifying their architectures using Tanh \cite{nwankpa2018activation} as the activation function. 
To mitigate errors, based on the analysis and the observations from \cite{liew2016bounded}, Hoang et al. \cite{hoang2020ft} presented a new version of the ReLU activation function to squash high-intensity activation values to zero, which is called Clip-Act. 
% It replaces the activation functions in the DNN with their clipped alternatives.
Ranger \cite{chen2020ranger} used value restriction in different DNN layers as a way to rectify the faulty outputs caused by transient faults.
An error correction technique based on the complementary robust layer (redundancy mechanism) and activation function modification (removal mechanism) has been recently introduced in \cite{ali2020erdnn}.
However, these methods, such as \cite{chen2020ranger,hong2019terminal,hoang2020ft,ali2020erdnn}, suffer from low fault coverage, since it is challenging to find a single global bound value for all neurons in a network layer. In Section~\ref{subsec:BoundRELIssu}, we will further demonstrate the limitations of such a globally bounded activation function used in \cite{chen2020ranger,hong2019terminal,hoang2020ft}. Motivated by this, in this paper, we introduce a fine-grained post-trainable activation function to improve the error resilience of DNNs under a wide range of fault rates.

%In this paper we solve this issue by introducing a new activation function bounding approach.
%\subsection{Our work}

\section{DNN Activation Function and Motivation}\label{sec:background}

%In this section, we briefly introduce the background of DNN models and their activation functions

\subsection{DNN Model and Activation Function}
Generally, we consider a DNN of depth $D$ corresponding to a neural network with an input layer, ${D-1}$ hidden layers, and an output layer. We denote the number of neurons as $N$. %at the $kth$ hidden layer
The  $k^{th}$ hidden layer receives an output $\bm{z}^{k-1}$ from the prior layer where an affine transformation is performed as the form of:
\equationskip
 \begin{equation}
   \mathscr{L}_{k}(\bm{z}^{k-1}) = \bm{w}^{k}\bm{z}^{k-1} + \bm{b}^{k}
 \end{equation}
where the network weights $\bm{w}^{k}$ and bias term $\bm{b}^{k}$ associated with the $k^{th}$ layer are chosen from independent and identically distributed samplings.% \cite{}.

A nonlinear activation function $\xi(.)$ is applied to each component of the transformed vector, before it is sent as an input to the next layer. The activation function is an identity function after an output layer.

Therefore, the final neural network representation is given by the composition:
\equationskip
 \begin{equation}
 \bm{u}_{\Theta}(\bm{z}) = (\mathscr{L}_{D} \circ\xi(\cdot) \circ \mathscr{L}_{D-1} \circ \xi(\cdot) \circ \dots \xi(\cdot) \circ \mathscr{L}_{1}) (\bm{z})
 \end{equation}
where the operator $\circ$ is the composition operator, $\Theta$ = ${\bm{w}^{k}, \bm{b}^{k}}:$ ${k=1,...,D}$, represents the trainable parameters in the network.
%,  and V is the parameter space; u is the output and x0 = x is the input.
%$\in$ $\scriptv$
%%%%%%%%%%%%%%%%%%%%
\subsection{ReLU Activation Function}
The choice of an activation function plays an important role in determining how a DNN learns and behaves. Many hand-engineered activation functions exist \cite{nwankpa2018activation}, but only a small number of them are widely used in modern DNN models. 
Among them, the Rectified Linear Unit (ReLU) function is the most widely used one in recent DNN models as it allows faster training convergence and less complex gradient computation \cite{krizhevsky2012imagenet}. The ReLU function, $\xi_{ReLU}(x)$, follows as \cite{arora2016understanding}: 
\equationskip
 \begin{equation}
     \xi_{ReLU}(x) = max(0,x)
 \end{equation}
 where $x$ is the input of the activation on all the input channels.
 
\subsection{Bounded ReLU: Remaining Issue and Motivation}
\label{subsec:BoundRELIssu}
%Although ReLU is now the de facto standard for activation functions of DNNs, many other functions are used that may be advantageous depending on the situation. 
Some recent studies \cite{chen2020ranger,hong2019terminal,hoang2020ft} revealed that bounding the output values of the activation function helps prevent the faults propagation in the network. In this regard, a  constrained range ReLU activation function could be deployed to enhance the resilience of DNN as \cite{chen2020ranger,hong2019terminal,hoang2020ft}:
\equationskip
 \begin{equation}
     \xi_{GBReLU}(x) =
     \begin{cases}
       0 & \text{if $x>\lambda$}\\
       x & \text{if $0<x\leq \lambda$}\\
       0 & \text{if $x\leq 0$}
     \end{cases}
 \end{equation}
where $\lambda$ is a bound value. It indicates that values greater than $\lambda$ are most likely faulty and need to be controlled by the activation function.
The value of $\lambda$ is determined by observing the maximum activation in all neurons of each layer \cite{chen2020ranger,hong2019terminal,hoang2020ft}; we call the approach used in \cite{chen2020ranger,hong2019terminal,hoang2020ft} as the globally bounded ReLU (GBReLU) and illustrate its limitation as below.

\begin{figure}[!tb]
 	\begin{center}
 	    \vspace{-0.1in}
 		\includegraphics[width = 0.45\textwidth]{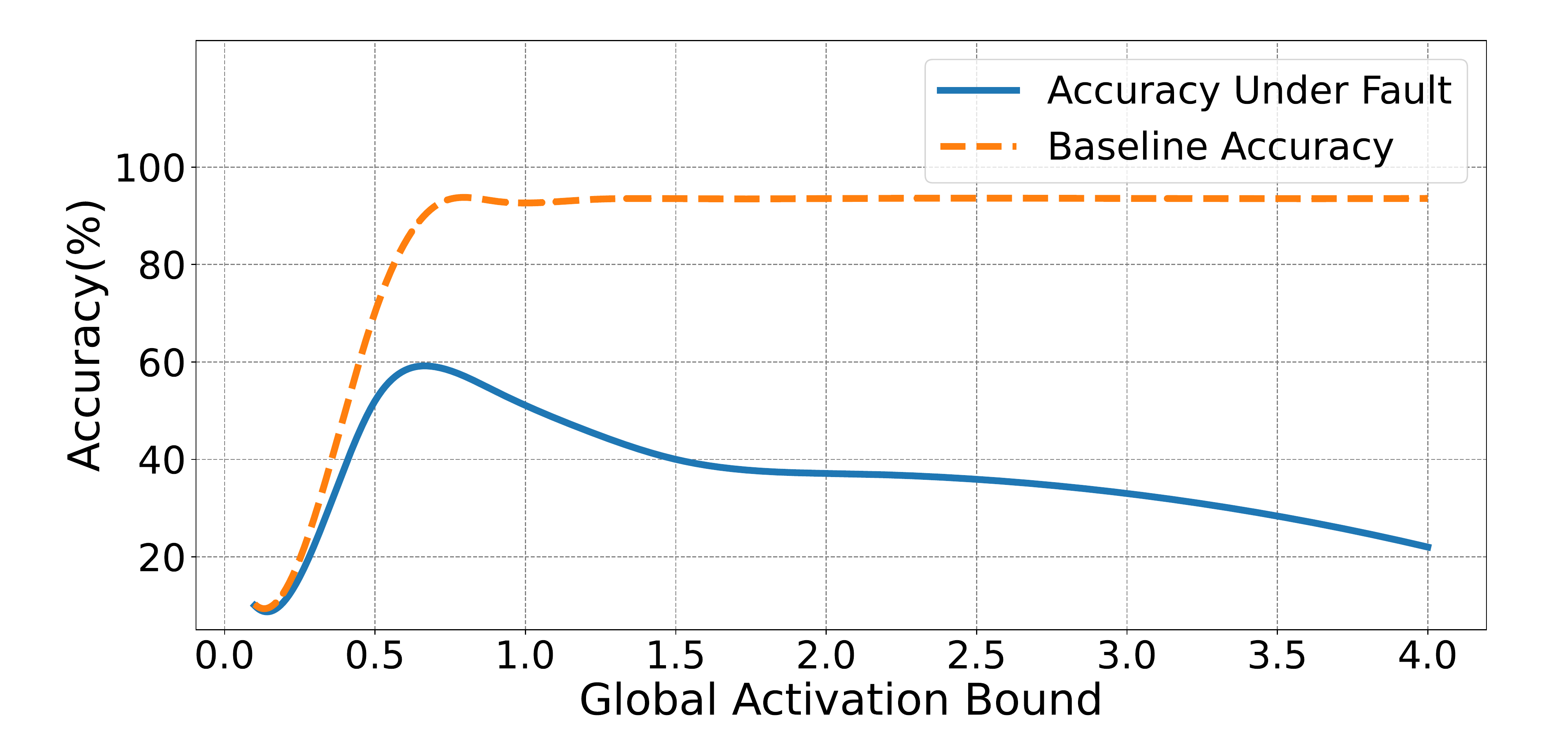}
 		\vspace{-0.05in}
 		\caption{Accuracy of the VGG16 network on CIFAR-10 dataset under a fault rate of $10^{-5}$, with regard to different global bound values of GBReLU in the second network layer.}
 		\vspace{-0.2in}
 		\label{fig:AvsR}
 	\end{center}	
 \end{figure}

%\subsubsection{Remaining Issue}
To evaluate the impact of GBReLU on the resilience of DNNs, we characterize the accuracy of the VGG16 network on the CIFAR-10 dataset under a fault rate of $10^{-5}$.
In this case study, we insert faults into the input layer and the second layer (convolutional layer) of the VGG16 network, and use GBReLU to replace the original ReLU in the second layer with a global bound value to restrict activation values of all neurons in this layer. The detailed experimental setup is in Section~\ref{sec:setup}. The results are shown in Figure~\ref{fig:AvsR}. 

First, there is a large accuracy gap between the baseline model without fault (called accuracy) and the model with GBReLU under a high fault rate of $10^{-5}$ (called resilience). Second, choosing a lower global activation bound ($\lambda$) could lead to a higher resilience (i.e., model accuracy under fault) until a certain threshold when the baseline accuracy without fault starts to drop significantly. More results on the accuracy of the GBReLU approach for different DNN models under different fault rates are presented in  Section~\ref{sec:resilience-results}. In general, when the fault rate goes higher beyond  $10^{-6}$, the GBReLU approach has a significant accuracy loss and is no longer effective in protecting the DNN model against faults.

% under fault using BReLU as a protection. 

% in the presence of faults via choosing the best $\lambda$ value. 
% As a result, deploying a global bound for all neurons (or all neurons in each layer) suffers from the accuracy loss of model. 

% We conduct experiments to   restricting activation by setting several bounds on BReLU. Our experimental results based on restricting activation magnitudes and fault-injection simulation are shown in Figure~\ref{fig:AvsR}. 

\begin{figure}[!tb]
 	\begin{center}
 	    \vspace{-0.1in}
 		\includegraphics[width = 0.45\textwidth]{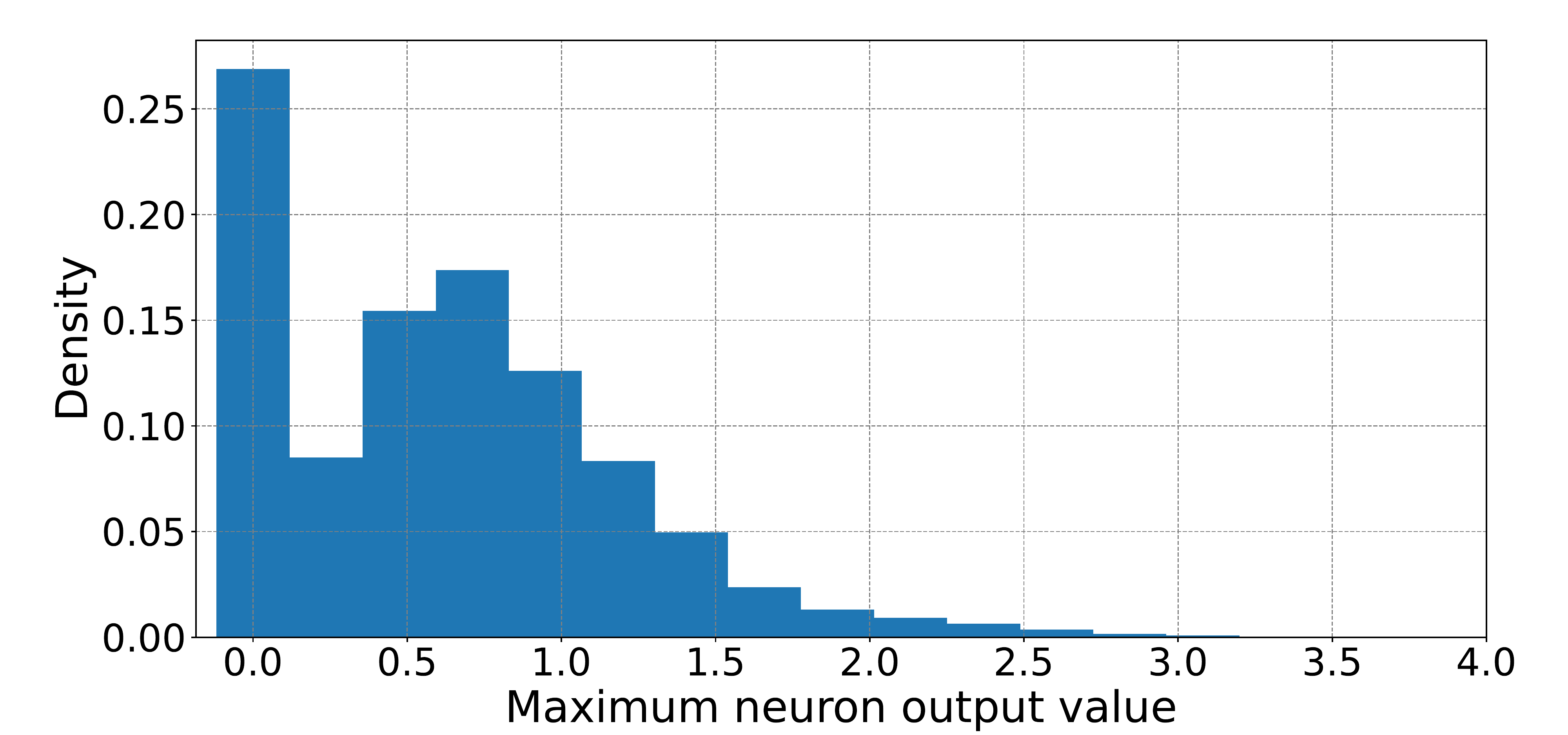}
 		\vspace{-0.05in}
 		\caption{Distribution of the maximum output values of all neurons in VGG16's second layer on the CIFAR-10 dataset.}
 		\vspace{-0.2in}
 		\label{fig:DisMaxNer}
 	\end{center}	
 \end{figure}

%\subsubsection{Motivation}
To better understand the source of this accuracy drop, we perform a statistical study on all neurons' output values in the second layer of the VGG16 network. Figure~\ref{fig:DisMaxNer} shows the distribution of the maximum output values for all neurons in the layer. We observe that the neurons across the layer have different maximum values; a similar trend is observed for other layers and other networks.
%and hence, we can conclude that the source of relative accuracy drop via bounding activation is mainly because of the fact that
However, the GBReLU approach \cite{chen2020ranger,hong2019terminal,hoang2020ft} chooses a fixed upper bound value for all neurons in the layer, which may not remove some faulty output values (if $\lambda$ is too big) or may eliminate some non-faulty output values (if $\lambda$ is too small). This motivates us to explore the fine-grained bounded activation function where we can set a bound for the activation function for each neuron in the network.

% Although, by selecting the smaller ($\lambda$), the robustness of DNN can be significantly increased, on the other hand, the accuracy of DNN may decrease as it may eliminate some non-faulty output values. In other worlds:
% \begin{itemize}
%     \item A conservative bounding approach is to set the bound to the large value such that it is less likely to affect the accuracy of the model.
%     \item Alternatively, one could choose a smaller bound to gain higher resilience boosting at the cost of accuracy loss.
% \end{itemize}

\section{FitReLU: Trainable Fine-Grained Bounded ReLU Activation Function}\label{sec:proposedAF} 
 
In light of our findings in Section~\ref{subsec:BoundRELIssu}, instead of defining a bounded activation function globally, we propose to define a fine-grained bounded activation function for each neuron.

\subsection{Neuron-Wise Bounded Activation Function}
Neuron-wise activation function acts as a vector of activation functions in each hidden-layer, where every neuron has its own activation function, as opposed to a single global activation function shared by all neurons.
In this paper, we introduce a parametric neuron-wise bounded ReLU (FitReLU-Naive) as:
\equationskip
 \begin{equation}
     \xi_{FitReLU-Naive}(x) =
     \begin{cases}
      0 & \text{if $x>\lambda_i$}\\
      x & \text{if $0<x\leq\lambda_i$}~~~~ \forall{i\in \{1,...,N\}} \\
      0 & \text{$x\leq 0$}
     \end{cases}
     \label{eq:NBRELU}
 \end{equation}
where $N$ denotes the number of neurons in the network and $\lambda_i$ denotes the bound value for neuron $i$. It indicates that values bigger than $\lambda_i$ are probably faulty for neuron $i$ and need to be controlled by its activation function. 
%Note that the activation can vary on different neurons and $\lambda_i$ is an adjustable parameter for each neuron $i$.
%, obtained during a post-training stage (section \ref{sec:proposedDNN}). %The motivation is intuitive and clear: 

\subsection{Issue with FitReLU-Naive}\label{subsubsec:Issu}
To deploy the proposed FitReLU-Naive in a DNN model, we need to adjust $N$ parameters, i.e., $\lambda_1$ to $\lambda_N$, so as to improve the resilience of the model. To find the best values of $\lambda_1$ to $\lambda_N$, we will introduce a post-training phase in Section \ref{sec:proposedDNN} to fine tune these values, in order to minimize the loss function and maximize the resilience. Most learning algorithms are based on the backward propagation of the error gradients. However, the derivative of the FitReLU-Naive activation function at point $x=\alpha$ cannot be computed, since it has different terms in each segment as presented in Equation~\ref{eq:NBRELU}. Therefore, it is nontrivial to learn the FitReLU-Naive activation functions (i.e., $\lambda_1$ to $\lambda_N$) via the post-training process.

\subsection{Trainable FitReLU}

\begin{figure}[!tb]
 	\begin{center}
 	    \vspace{-0.05in}
 		\includegraphics[width = 0.47\textwidth]{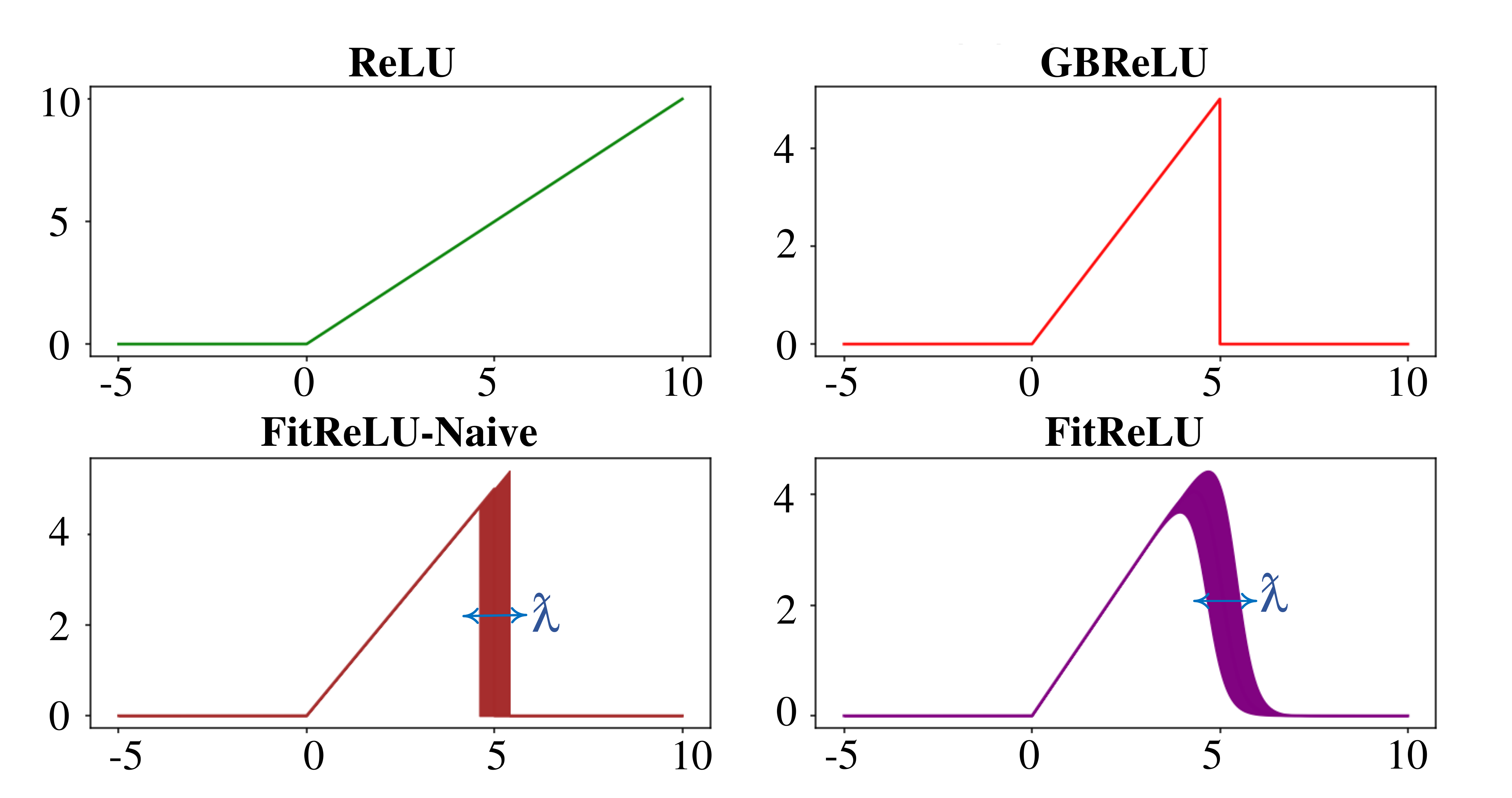}
 		\vspace{-0.1in}
 		\caption{Representation of original ReLU \cite{arora2016understanding}, GBReLU (globally bounded ReLU) \cite{hoang2020ft}, naive FitReLU and trainable FitReLU activation functions.}
 		\vspace{-0.3in}
 		\label{fig:TNBRELU}
 	\end{center}	
 \end{figure}

To solve the above problem in Section~\ref{subsubsec:Issu}, we introduce a new activation function (called FitReLU) that can be trained during the learning process \cite{apicella2021survey} and has a behavior similar to FitReLU-Naive. Inspired by the sigmoid mathematical function \cite{nwankpa2018activation}, we reformulate FitReLU into a trainable one as:

\equationskip
\begin{equation}
\xi_{FitReLU}(x) = max (0,\ x -  \frac{x}{1 + e^{k(x - \lambda_i )}}) ~~~\forall{i\in \{1,...,N\}}
     \label{eq:TNBRELU}
\end{equation}
where $N$ and $\lambda_i$ are the same as those in Equation~\ref{eq:NBRELU}, and $k$ is a coefficient to control the decent slope of the function and is empirically computed.
 The illustrations of both FitReLU-Naive and trainable FitReLU, as well as the original ReLU and GBReLU are presented in Figure~\ref{fig:TNBRELU}.
 
\section{\Toolname: The Proposed Resilient DNN Framework}\label{sec:proposedDNN}

\begin{figure}[!tb]
 	\begin{center}
 	    \vspace{-0.1in}
 		\includegraphics[width = 0.4\textwidth]{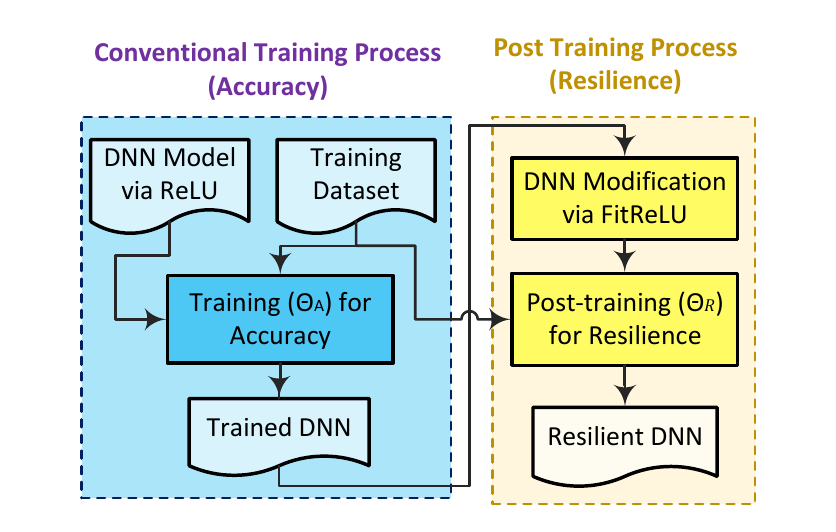}
 		\vspace{-0.1in}
 		\caption{\Toolname: our two-stage workflow to enhance the resilience of DNNs by post-training FitReLU activation functions.}
 		\vspace{-0.35in}
 		\label{fig:DNN}
 	\end{center}	
 \end{figure}

%The expansion of the DNN parameter space generates a high-dimensional optimization problem, the solution to which can be difficult to find. Hence, in order 
To avoid retraining the DNN model from scratch and complicating the DNN training process, we divide our workflow (called \Toolname) of building an error resilient DNN with the proposed FitReLU activation function into two separate stages.
As shown in Figure~\ref{fig:DNN}, in the first stage, \Toolname~trains a DNN model with the original ReLU activation function using the conventional training process. Its goal is to learn the weight and bias parameters ($\bm{w}^{1}$ to $\bm{w}^{D}$, $\bm{b}^{1}$ to $\bm{b}^{D}$) to improve the model accuracy, without the consideration of error resilience. In the second stage, \Toolname~first replaces the ReLU activation function in the trained DNN model with our proposed FitReLU variants. Then it post-trains the modified DNN model to tune the fine-grained bound values ($\lambda_1$ to $\lambda_N$) of FitReLU functions to improve the model resilience against faults.
%Such a separation avoids the need to retrain the DNN model from scratch to enhance the model resilience.

Formally, the final error resilient DNN model can be represented using two separable groups of learning parameters as:
\equationskip
 \begin{equation}
 \bm{u}_{\Theta_{A},\Theta_{R}}(\bm{z}) = (\mathscr{L}_{D} \circ \xi_{D-1}(\cdot) \circ \mathscr{L}_{D-1} \circ \xi_{D-2}(\cdot) \circ \dots \xi_{1}(\cdot) \circ \mathscr{L}_{1}) (\bm{z})
 \end{equation}
where the set of trainable parameters $\Theta_{A}$ consists of ${\bm{w}^{k}, \bm{b}^{k}}:$ ${k=1,...,D}$, which indicate the accuracy goal, and $\Theta_{R}$ denotes the post-trainable parameters, i.e., ${\lambda_i}: {i=1,...,N}$, which indicate the resilience goal.

%\subsection{DNN Architecture Model}

\noindent \textbf{Conventional Training for Accuracy:}
A typical learning algorithm for the training stage updates $\Theta_{A}$  through gradient computations, and calculates the errors on training dataset $\mathbb{D}$ via an accuracy loss function $\mathcal{L}(\mathbb{D};\Theta_{A})$. We denote the accuracy of the trained network as $\mathcal{A}(\Theta_{A})$.

\noindent \textbf{DNN Architecture Modification:}
We modify the trained DNN model by replacing the original ReLU function for each neuron with our FitReLU variants, and initialize the bound parameters $\Theta_{R}$ for each neuron to their maximum values over the training dataset $\mathbb{D}$. Note that by applying this modification, the accuracy of the obtained model, $\mathcal{A}(\Theta_{A},\Theta_{R})$, would be changed.

\noindent \textbf{Post-Training for Resilience Enhancement:}
To enhance the error resilience of the modified DNN architecture, we perform a post-training phase detailed in the following subsections.

\subsection{Resilience Enhancement as an Optimization Problem}

The main goals of our post-training phase could be expressed as: 1) maximizing the resilience of the model, and 2) maintaining the accuracy of the model. Using goal 1 as the objective and goal 2 as the constraint, we can formulate the model resilience enhancement as the following optimization problem:
\equationskip
 \begin{equation}
 \begin{split}
     &max_{~\bm{{\Theta_{R}}}} (\mathcal{R}(\Theta_{A},\Theta_{R}))\\
     & s.t. \quad \quad \mathcal{A}(\Theta_{A})- \mathcal{A}(\Theta_{A},\Theta_{R})<\delta
 \end{split}
 \label{eq:opt1}
 \end{equation}
where $\mathcal{R}(\Theta_{A},\Theta_{R})$ is the resilience of the model, which represents the accuracy of the modified network under faults. $\mathcal{A}(\Theta_{A},\Theta_{R})$ is the accuracy of the modified network without faults. And the threshold $\delta$ can be adjusted based on how much accuracy loss we are willing to accept for resilience boosting. 

As discussed in Section~\ref{subsec:BoundRELIssu}, having lower values of $\Theta_{R}$ could increase the chance of fault removing and thus increase the resilience. However, when the values of $\Theta_{R}$ become lower than a certain level, the accuracy loss could also increase. Therefore, we can maximize the resilience by minimizing $\Theta_{R}$ values until the accuracy loss is not acceptable. The optimization problem in Equation~\ref{eq:opt1} can be reformulated as:
\equationskip
 \begin{equation}\label{optimization}
 \begin{split}
     min ({\Theta_{R}}) \quad \quad s.t. \ \  \mathcal{A}(\Theta_{A})- \mathcal{A}(\Theta_{A},\Theta_{R})<\delta
 \end{split}
 \end{equation}
 
\subsection{Post-Training Algorithm}
In order to minimize $\Theta_{R}$ while maintaining the accuracy (i.e., Equation~\ref{optimization}), we introduce a post-training algorithm with a new loss function $\mathcal{L}(\mathbb{D};\Theta_{A}, \Theta_{R}$). It is the same as that in the conventional training step except that it has an extra regularization restriction over the $\Theta_{R}$ parameters to penalize the increasing upper bounds. That is:
\equationskip
\begin{equation}
    \mathcal{L}(\mathbb{D};\Theta_{A}, \Theta_{R}) =  \mathcal{L}(\mathbb{D};\Theta_{A}) + \frac{\zeta}{N} \sum_{i=1}^{N} \lambda_i^2
\end{equation}
where 
%$N$ is the number of neurons in all over the network and 
$\zeta$ is a hyper parameter. Note that only bound values $\Theta_{R}$ would be adjusted and none of $\Theta_{A}$ will be changed in this step.

The solution to this problem can be approximated using an iterative gradient descent algorithm. In this work, we use the ADAM optimizer \cite{kingma2014adam} to solve it, which is a widely used variant of the stochastic gradient descent algorithm.
\section{Experimental Results}\label{Result}
%We conducted some experiments to evaluate the efficiency of our proposed approach.

\subsection{Experimental Setup} \label{sec:setup}
\subsubsection{DNN Model and Dataset}
We test \Toolname~on three widely used DNN models: AlexNet, VGG16 and ResNet50. We use 32-bit fixed-point representation (1 sign bit, 15 integeral bits and 16 fractional bits) rather than floating-point for model parameters, since it is more energy efficient. 
These models are trained on two different datasets: CIFAR-10 and CIFAR-100. 
All experiments are performed on an Intel Core i7@3.2 GHz processor with 32 GB memory and an NVIDIA TITAN V GPU.

%CIFAR-100: vgg:73.67  ResNet:78.43  AlexNet: 60.53
%CIFAR-10: vgg:93.57   ResNet:95.03  AlexNet:87.08

To validate the efficacy of our approach, we compute the top-1 classification accuracy. 
The baseline (top-1) accuracy values without fault for CIFAR-10 dataset are 95.03\%, 93.57\% and 87.08\%, for ResNet50, VGG16 and AlexNet models, respectively.
The baseline accuracy values of CIFAR-100 dataset are 78.43\%, 73.67\% and 87.08\% for the three models, respectively.

% \begin{table}[tb!]
% \begin{center}
% \vspace{0.05in}
% \caption{Base accuracy of DNN models}
% \vspace{-0.05in}   
% \begin{tabular}{|c| c| c|} 
%  \hline
%  Network & CIFAR-10 Acc(\%) & CIFAR-100 Acc(\%) \\ [0.5ex] 
%  \hline\hline
%  ResNet-50 & 95.03 & 78.4 \\ [0.5ex] 
%  \hline
%  VGG16 & 93.57 & 73.6 \\ [0.5ex] 
%  \hline
%  AlexNet & 87.08 & 60.53 \\ [0.5ex] 
%  \hline
% \end{tabular}
%     %\vspace{-0.05in}
% \label{tab:acc}
% \end{center}
% \vspace{-0.15in}
% \end{table}

% \begin{center}
%  \begin{tabular}{|c| c| c|} 
%  \hline
%  Network & CIFAR-10 Acc(\%) & CIFAR-100 Acc(\%) \\ [0.5ex] 
%  \hline\hline
%  ResNet-50 & 95.03 & 78.4 \\ [0.5ex] 
%  \hline
%  VGG16 & 93.57 & 73.6 \\ [0.5ex] 
%  \hline
%  AlexNet & 87.08 & 60.53 \\ [0.5ex] 
%  \hline
% \end{tabular}
% \label{tab:acc}
% \end{center}

\subsubsection{Fault Simulation}
In this paper, we consider memory faults to analyze the error resilience of DNNs, where the model parameters may be affected under the injection of random bit-flips in the memory blocks that store these parameters. The weights and biases of different layers, as well as parameters of activation functions, are considered as the fault space.
To simulate random fault injections, we develop a fault injection tool based on the PyTorch framework. 
We assume that the fault space would be distributed uniformly over random locations in the target units and throughout the inference period, which is in line with prior studies \cite{reagen2018ares,hoang2020ft,chen2020ranger}. We perform fault injections on different fault rates ranging from $10^{-7}$ to $3\times10^{-5}$.

\subsection{Resilience Results} \label{sec:resilience-results}

To demonstrate the efficacy of \Toolname, we compare the model resilience under different fault rates, between our \Toolname, state-of-the-art studies Clip-Act \cite{hoang2020ft} and Ranger \cite{chen2020ranger} that use a layer-wise globally bounded ReLU activation function to improve model resilience,
%(\zf{provides more details about Ranger and confirms the one for Clip-Act}), 
as well as unprotected DNNs.

\begin{figure}[!tb]
	\begin{center}
	    \vspace{-0.1in}
 		\includegraphics[width=0.5\textwidth]{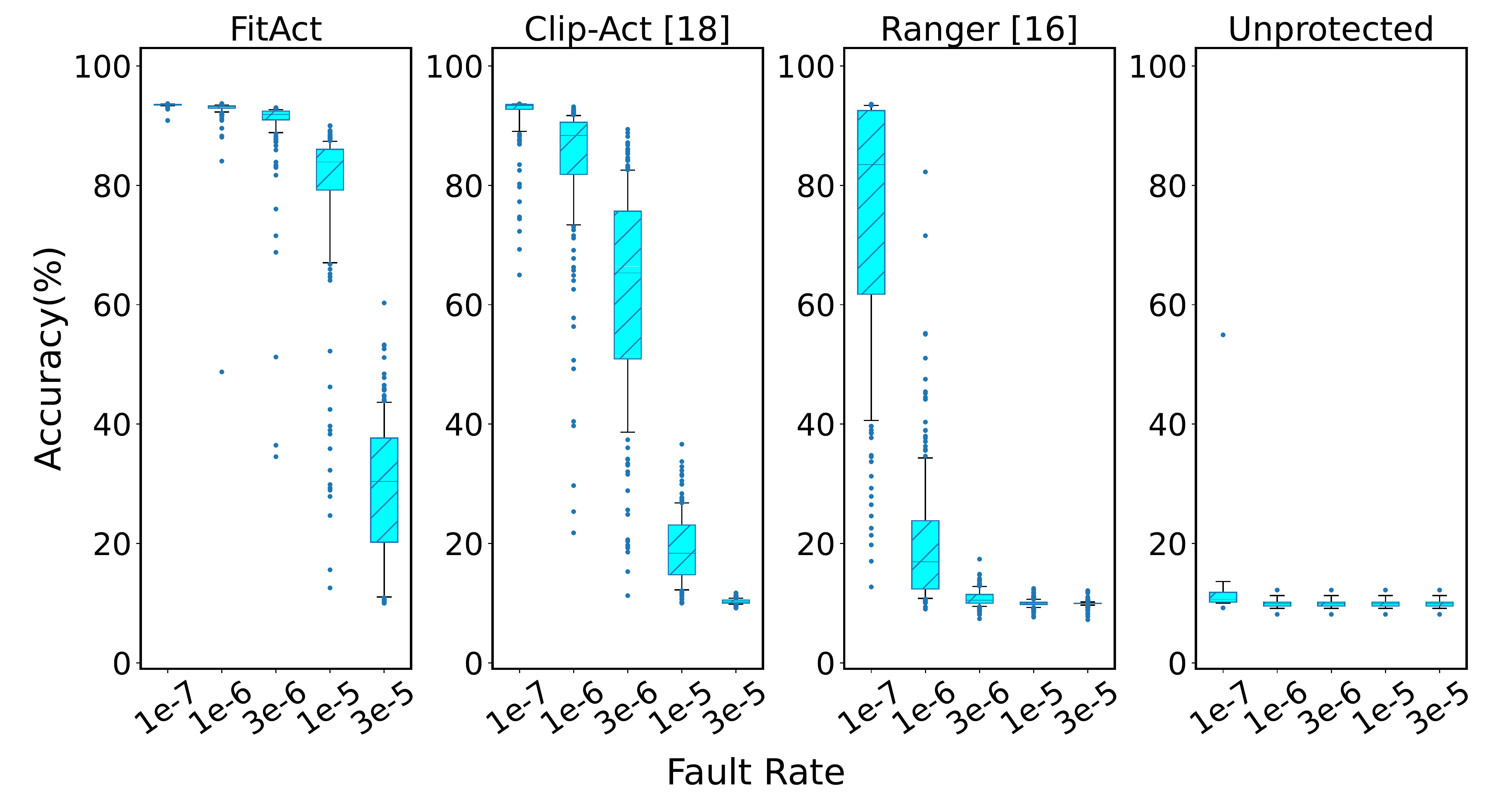}
 	\vspace{-0.2in}
	\caption{Model accuracy distribution for \Toolname, Clip-Act \cite{hoang2020ft}, Ranger \cite{chen2020ranger} and unprotected DNN,
	using the VGG16 network on CIFAR-10 dataset, under different fault rates.}
	\vspace{-0.3 in}
	\label{fig:distvgg}
	\end{center}	
\end{figure}

Figure~\ref{fig:distvgg} compares the model accuracy distribution between different approaches under different fault rates, using the VGG16 network on CIFAR-10 dataset as a case study. The accuracy of the unprotected DNN model has a huge drop to around 10\% under fault injection, which clearly calls for resilience enhancement. In general, as the fault rate increases, the accuracy of the three protected DNNs decreases, because the protection becomes less effective and faults are more likely to be injected in critical bits, having a more substantial impact on the output classification. As shown in Figure~\ref{fig:distvgg},  \Toolname~can keep the accuracy within an acceptable range even at a high fault rate of $10^{-5}$, whereas Clip-Act \cite{hoang2020ft} and Ranger \cite{chen2020ranger} experience a significant accuracy drop at fault rates higher than $10^{-6}$ and $10^{-7}$, respectively.

\begin{figure*}[!ht]
	\begin{center}
		    \vspace{-0.33in}
		\includegraphics[scale=0.34]{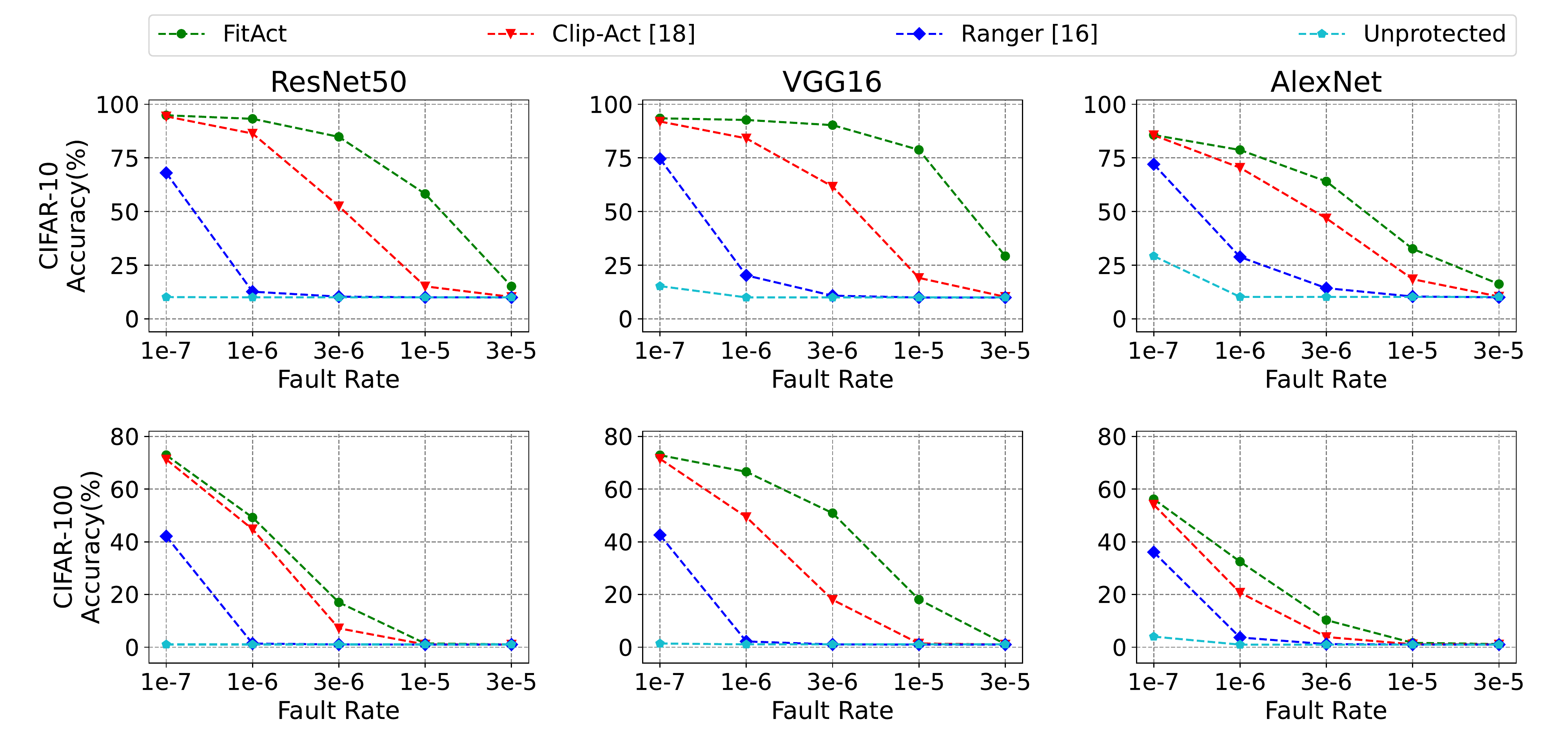}
		\vspace{-0.1in}
		\caption{Average model accuracy for \Toolname, Clip-Act \cite{hoang2020ft}, Ranger \cite{chen2020ranger} and 
		unprotected DNN models, including ResNet50, VGG16, and AlexNet models on CIFAR-10 and CAFIR-100 datasets, under different fault rates.}
		\label{fig:aveAcc}
		\vspace{-0.3in}
	\end{center}	
\end{figure*}

Figure~\ref{fig:aveAcc} compares the average accuracy between \Toolname, Clip-Act \cite{hoang2020ft}, Ranger \cite{chen2020ranger} and unprotected model under different fault rates, for various DNN architectures and datasets. 
First, all three protection methods boost the accuracy under fault compared to the unprotected one, and \Toolname~achieves the best accuracy (i.e., resilience) among all.
Second, compared to Clip-Act \cite{hoang2020ft}, when the fault rate becomes $3\times10^{-6}$ or higher, \Toolname~achieves a significantly higher accuracy. For example, at the fault rate of $3\times10^{-6}$, \Toolname~can still achieve a high accuracy of 84.81\% and 90.28\% for ResNet50 and VGG16 models on CIFAR-10 dataset, while Clip-Act can only achieve an accuracy of 52.47\% and 61.63\%. 
Third, Ranger \cite{chen2020ranger} performs much worse than Clip-Act \cite{hoang2020ft} and \Toolname, even at a low fault rate of $10^{-7}$. At a higher fault rate than $10^{-7}$, Ranger is no longer effective in the protection against faults.
This is because Ranger truncates an output faulty value to a big positive bound, which still propagates in the network.

\subsection{Time and Memory Space Overhead}

\subsubsection{Training Step}
%               ResNet  vgg AlexNet
% training      340 ,    60      and 17
% posttrain    21,       4,       and  1
% overhead   6.2%,   6.7%   and 5.9%
Since \Toolname~separates the conventional training for accuracy and post-training for resilience, it can maintain a small computational overhead. For example, 
%Compared to traditional activation function, the locally adaptive activation function-based DNN has additional parameters to adjust in post train step. Thus, we explore the additional computational cost required in post-training step.
the whole process of post-training for ResNet50, VGG16 and AlexNet models on CIFAR-10 dataset take about 21, 4 and 1 minutes, respectively. Considering the long runtime for conventional training of these models (340, 60, and 17 minutes), \Toolname~only adds around 5.9\% to 6.7\% extra runtime overhead.

\begin{table}[tb!]
\begin{center}
\vspace{0.1in}
\caption{Runtime and memory space overheads (O/H) when deploying resilient DNN with \Toolname~in the inference stage.}
\vspace{-0.05in}    
\scalebox{0.96}{
\begin{tabular}
{|c|l|r|r|r|r|r|r|}
    \hline 

    \multicolumn{2}{|c|}{} &  \multicolumn{3}{|c|}{Runtime (ms)}  & \multicolumn{3}{|c|}{Memory (Mb)} \\   \cline{3-8}
    \multicolumn{2}{|c|}{} & ReLU & \Toolname & O/H  & ReLU & \Toolname & O/H\rule{0pt}{10pt} \\   \hline \hline
    \multirow{3}{*}{\rotatebox{90}{CIFAR-10}} & ResNet50 & 9.820&10.869&10.68\% & 156.69 & 165.05 & 5.34\%\rule{0pt}{10pt}\\  
    \cline{2-8}  & VGG16 & 5.324&5.600&5.18\% & 29.60 & 30.46 & 2.91\%\rule{0pt}{10pt}\\  
    \cline{2-8}  & AlexNet & 3.037&3.175&4.54\% & 23.53 & 23.68 & 0.64\%\rule{0pt}{10pt}  \\
    \hline
    \multirow{3}{*}{\rotatebox[origin=c]{90}{CIFAR-100}} & ResNet50 & 9.877&10.974&11.11\% & 156.70 & 165.10 & 5.36\%\rule{0pt}{11pt}
    \\  \cline{2-8}  & VGG16 & 5.178&5.487&5.97\% & 29.65 & 30.47 & 2.77\%\rule{0pt}{11pt}
     \\  \cline{2-8}  & AlexNet & 3.001 & 3.224 & 7.43 \% & 23.53 & 23.69 & 0.68\%\rule{0pt}{11.5pt} \\
    \hline
    
\end{tabular}
}
%\vspace{-0.05in}
    \label{tab:overhead}
\end{center}
\vspace{-0.4in}
\end{table}

\subsubsection{Inference Step}
% Since an activation function is applied to the outputs of most DNN neurons, it will mainly contribute to the overall DNN inference time and memory usage footprint.
% The proposed \Toolname~adds an extra computation to the inference step with FitReLU which lead to memory and runtime overheads via inference step.

Compared to the original ReLU, FitReLU (Equation~\ref{eq:TNBRELU}) costs more computation and needs extra memory space to store the bound values $\lambda_1$ to $\lambda_N$.
Table~\ref{tab:overhead} presents the time and memory space overheads of \Toolname~for various models and datasets in the inference step. Since most of the computations and memory storage happen in the convolutional and fully connected layers, the overheads introduced by \Toolname~is manageable. Shown in Table~\ref{tab:overhead}, the runtime overheads are less than 12\% and the memory space overheads are less than 6\%.
\section{Conclusion}\label{conclusion}

% In this paper, we propose \Toolname, a low-cost approach to enhance the error resilience of DNNs by deploying fine-grained post-trainable activation functions. 
% The main idea is to precisely bound the activation value of each individual neuron via neuron-wise bounded activation functions, so that it could prevent the fault propagation in the network. 
% To avoid complex DNN model re-training, we propose to decouple the accuracy training and resilience training, and develop a lightweight post-training phase to learn these activation functions with precise bound values.
% Experimental results on widely used DNN models such as AlexNet, VGG16, and ResNet50 demonstrate that \Toolname~outperform state-of-the-art studies such as Clip-Act and Ranger in enhancing the DNN error resilience for a wide range of fault rates, while adding manageable runtime and memory space overheads.

In this paper, we have proposed \Toolname, a framework to build low-overhead error-resilient DNNs by deploying fine-grained post-trainable activation functions. First, we have proposed a fine-grained, trainable, neuron-wise activation function called FitReLU to precisely bound the activation value of each neuron, so as to prevent the fault propagation in the network.
Second, we have developed a lightweight post-training phase to learn all these activation functions with precise bound values, which is decoupled from the conventional training for model accuracy. Finally, we have conducted a wide range of experiments using fault simulation
% , compared to state-of-the-art studies Clip-Act \cite{hoang2020ft} and Ranger \cite{chen2020ranger}, 
and confirmed the effectiveness and advantages of our \Toolname~framework in enhancing the error resilience for widely used DNN models.

% modifying the network architecture to create error-resilient DNNs. We demonstrated that establishing upper limits on activation functions for each neuron can help to more precisely prevent fault propagating to subsequent DNN layers thorough an in-depth investigation of the values of neurons. In this regard, we introduced a novel fine-grained trainable activation function to further improve the resilience of network. We also proposed a gradient-ascent based search for adjusting the boundaries of neuron wised activations. 
% Our experiments confirmed that our proposed activation function significantly improves the error-resiliency of DNNs.
% Thanks to our \Toolname~framework, it is now possible to deploy highly resilient DNNs automatically in resource-constrained devices.
\section*{Acknowledgements}
We acknowledge the support from Government of Canada Technology Demonstration Program and MDA Systems Ltd; NSERC Discovery Grant RGPIN341516, RGPIN-2019-04613, DGECR-2019-00120, Alliance Grant ALLRP-552042-2020, COHESA (NETGP485577-15), CWSE PDF (470957); CFI John R. Evans Leaders Fund; Simon Fraser University New Faculty Start-up Grant.
% NSERC Discovery Grant RGPIN-2019-04613 and DGECR-2019-00120; Canada Foundation for Innovation John R. Evans Leaders Fund;

\bibliographystyle{IEEEtran}
\bibliography{references}

% \bibliography{references}	
% \end{thebibliography}

% \bibliographystyle{ACM-Reference-Format}

\end{document}